\documentclass[nohyperref]{article}

\usepackage{microtype}
\usepackage{graphicx}
\usepackage{subfigure}
\usepackage{booktabs} 
\usepackage{mathtools, nccmath}
\newcommand{\round}[1]{\ensuremath{\lfloor#1\rceil}}

\usepackage{hyperref}

\usepackage[accepted]{icml2022}

\usepackage{amsmath}
\usepackage{amssymb}
\usepackage{mathtools}
\usepackage{amsthm}

\usepackage[capitalize,noabbrev]{cleveref}

\theoremstyle{plain}

\theoremstyle{definition}

\theoremstyle{remark}

\DeclareMathOperator*{\argmax}{arg\,max}

\usepackage[textsize=tiny]{todonotes}

\icmltitlerunning{Diverse Concept Proposals for Concept Bottleneck Models}

\begin{document}

\twocolumn[
\icmltitle{Diverse Concept Proposals for Concept Bottleneck Models}




\begin{icmlauthorlist}
\icmlauthor{Katrina Brown}{seas}
\icmlauthor{Marton Havasi}{seas}
\icmlauthor{Finale Doshi-Velez}{seas}
\end{icmlauthorlist}

\icmlaffiliation{seas}{Harvard SEAS, Cambridge MA, USA}

\icmlcorrespondingauthor{Katrina Brown}{katrinabrown@college.harvard.edu}

\icmlkeywords{Concept bottleneck models, diverse explanations}

\vskip 0.3in
]



\printAffiliationsAndNotice{}  

\begin{abstract}
Concept bottleneck models are interpretable predictive models that are often used in domains where model trust is a key priority, such as healthcare. They identify a small number of human-interpretable concepts in the data, which they then use to make predictions.
Learning relevant concepts from data proves to be a challenging task.
The most predictive concepts may not align with expert intuition, thus, failing interpretability with no recourse.
Our proposed approach identifies a number of predictive concepts that explain the data. By offering multiple alternative explanations, we allow the human expert to choose the one that best aligns with their expectation.
To demonstrate our method, we show that it is able discover all possible concept representations on a synthetic dataset. On EHR data, our model was able to identify 4 out of the 5 pre-defined concepts without supervision.
\end{abstract}

\section{Introduction}
\label{introduction}
Recently popularized by \citet{koh2020concept}, concept bottleneck models (CBM) prove to be a highly flexible and (hopefully) interpretable model class.  Inputs $x$ are first converted to a small set of concepts $c$ that are used to make the ultimate prediction $y$. This allows the human operator to inspect the concepts that form the basis of the prediction and make an informed decision on whether to trust the prediction or not.

For the CBM to be effective, the concepts $c$ must make semantic sense to the human user. Unfortunately, many datasets do not come with ground truth concept labels, and there is no statistical reason that learning concepts that are highly predictive from data will result in concepts that have semantic meaning \citep{caruana2015intelligible}.  However, there may exist an alternate set of distinct concepts that is comparable in predictive performance but is more interpretable and thus preferred by human users. Existing methods do not discuss how to identify these alternate, preferable concepts. 

In this work, we propose a method for generating multiple sets of possible concepts that can be offered to a human expert to choose from.  The method works by first generating a large number of plausible concepts, which are then trimmed to only include the most predictive ones. These concepts, however, contain highly correlated ones. Therefore, we select a diverse set of proposals from the remaining concepts. We compare and contrast methods different methods and diversity metrics for constructing this final set.



Our contributions are the following: 
\begin{itemize}
  \item We introduce a method for identifying a small, diverse set of concept models for expert inspection.
  \item Our approach can straightforwardly generate concepts that complete known ones (i.e. if some concept labels are provided, or an expert finds some of the discovered concepts meaningful and others not).
  \item On a 2D synthetic dataset that can be explained by more than 7 meaningfully different concept combinations, we find 6; a naive method finds only 1.
  \item On the real MIMIC-III EHR dataset, our method discovers 4 of the 5 synthetic ground truth concepts.
\end{itemize}




\section{Related Works}
\label{related}

Proposing multiple explanations has been studied in problem specific settings e.g. \citet{liu2017multiple}. Closer to our work, \citet{ross2018learning} attempts to capture all distinct explanations for a dataset. Notably different from our work, they find the maximal set of explanations for rule-based models as opposed to finding a restricted set of explanations for concept bottleneck models.

Related methods for concept discovery include concept whitening models \citep{chen2020concept} that  use concept activation vector methods to analyse networks that have been pre-trained. \citet{yeh2019completeness} attempts to propose concept-based explanations that satisfy the completeness requirement, meaning that they fully explain the predictions of the underlying model. Neither of these works is able to propose multiple explanations in case the initial explanation is not semantically meaningful.


\section{Methods}
\label{methods}
\begin{figure}[h]
	\centering
	\includegraphics[width=0.7\columnwidth]{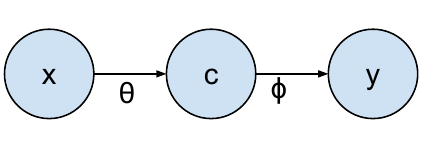}
	\caption{Graphical model of a Concept Bottleneck Model (CBM). The inputs $x$ are used to predict the distribution of concepts $p_\theta(c|x)$ (parameterized by $\theta$). Then, the concepts are used to predict the final label $p_\phi(y|c)$ (parameterized by $\phi$).}
	\label{fig:naive}
\end{figure}
Denote the input-label pairs in the dataset by $\{(\mathbf{x}^{(n)}, \mathbf{y}^{(n)})\}_{n=1}^N$.  A concept bottleneck model has the architecture shown in Figure~\ref{fig:naive}.  Our goal is to find concepts $\{\mathbf{c}^{(n)}\}_{n=1}^N$ (and parameters $\theta$ and $\phi$) that are both predictive of $\mathbf{y}$ and interpretable. 

Our method computes a set of $M$ concept proposals that are all predictive that we can present to a human expert to select the most interpretable one (or interpretable concepts).  We find these $M$ proposals in a two-stage approach.  First, we draw a large number of samples from the posterior $p(\mathbf{c}, \theta, \phi |\mathbf{x}, \mathbf{y})$ (we assume that the prior distributions for the concept predictor and label predictor are provided as $p(\theta)$ and $p(\phi)$ respectively) using Hamiltonian Markov Chain Monte Carlo (HMC). 

Second, we trim down these proposals to a small, predictive, and diverse subset. We ensure that we only consider predictive $(\theta, \phi)$ pairs by filtering out the ones below a baseline level of accuracy $t_{acc}$.  Let $\mathcal{C}$ be the set of proposals with predictive accuracy at least $t_{acc}$ on the dataset.  Many of the proposals in $\mathcal{C}$ will be highly correlated.  Below, we describe several approaches for identifying diverse sets of proposals.


\subsection{Constructing a diverse set of concept set proposals}
\label{sec:diverse}

In our results, we compare two algorithms (greedy construction and K-means clustering) for selecting the diverse subset. Each of these rely on a similarity metric $\mathcal{D}(\mathbf{\hat{c}}, \mathbf{\tilde{c}})$ that allow them to select diverse proposals.  

\subsubsection{Choices for Similarity Metrics}
We consider four similarity metrics commonly used in the literature:
\begin{enumerate}
    \item Euclidean Distance
    \begin{equation}
    \mathcal{D}_{euc}(\boldsymbol{\hat{c}},\boldsymbol{\Tilde{c}})= \sqrt{\sum \limits_{n=1}^N ||\boldsymbol{\hat{c}}^{(n)}-\boldsymbol{\Tilde{c}}^{(n)}||^2}
    \end{equation}
    \item Cosine Similarity
    \begin{equation}
    \mathcal{D}_{cos}(\boldsymbol{\hat{c}},\boldsymbol{\Tilde{c}}) = \frac{\boldsymbol{\hat{c}}\cdot \boldsymbol{\Tilde{c}}}{||\boldsymbol{\hat{c}}||\cdot||\boldsymbol{\Tilde{c}}||}
    \end{equation}
    \item Absolute
    \begin{equation}
    \mathcal{D}_{abs}(\boldsymbol{\hat{c}},\boldsymbol{\Tilde{c}})=\sum \limits_{n=1}^N||\boldsymbol{\hat{c}}^{(n)}-\boldsymbol{\Tilde{c}}^{(n)}||_1
    \end{equation}
    \item Percent disagreement
    \begin{equation}
    \mathcal{D}_{pct}(\boldsymbol{\hat{c}},\boldsymbol{\Tilde{c}})=\sum \limits_{n=1}^N||\round{\boldsymbol{\hat{c}}^{(n)}}-\round{\boldsymbol{\Tilde{c}}^{(n)}}||_1
    \end{equation}
\end{enumerate}

\subsubsection{Selecting the Diverse Subset}
We compare two options for constructing the diverse subset.

\paragraph{Greedy construction}
With our distance metric, we can greedy construct a subset $\mathcal{P}\subset \mathcal{C}$ (of size $M$) of diverse proposals. Greedy construction starts with the subset containing a random concept proposal $\mathcal{P}=\{\mathbf{c}\}$. Then, over $M-1$ iterations, we greedily extend this subset to the target size. In each iteration, we select $\mathbf{\hat{c}}$ such that the most similar concept in $\mathcal{P}$ is maximally different (according to one of our distance metrics above):
\begin{equation}
    \mathcal{P} \leftarrow \mathcal{P} \cup \{\underset{\mathbf{\hat{c}} \in \mathcal{C}}{\argmax}[\min \{\mathcal{D}(\mathbf{\hat{c}}, \mathbf{\tilde{c}})|\mathbf{\tilde{c}} \in \mathcal{P}\}]\}
\end{equation}


\paragraph{K-means clustering}
Alternatively to the greedy construction, we can construct the proposal set by grouping the similar proposals into $M$ clusters. For this, we use classical K-means clustering with either Euclidean distance or Cosine similarity (Absolute and Percent disagreement are not suitable for K-means).
Once the $M$ clusters are determined, we select the concept proposals closest to the cluster means.




\subsection{Proposing and Conditioning on Individual Concepts}
The approach above found entire models---collections of concepts--that were collectively predictive.  Another option is to present a collection of individual concepts from which the human expert can choose.

Creating a diverse collection of individual concepts is analogous to constructing a diverse set of concept sets.  The only difference is that once we obtain the large set of concept samples $\mathcal{C}$, we split these into individual concepts before continuing to select a small, diverse subset.

However, ideally we would want to provide the expert guidance on they might need to augment their selected concepts to get good predictive performance: once an interpretable concept $c_i$ is identified by the expert, we want to condition on it and propose further concepts that work well with it. To do this, we simply run our algorithm for generating proposals, but sample the proposals conditional on $c_i$: $p(\mathbf{c}|\mathbf{x}, \mathbf{y}, c_i)$.  The expert can continue selecting concepts until the full model reaches a desired level of interpretability and accuracy.

\section{Experimental Setup}
We start by demonstrating our method on a synthetic dataset that admits a large number of concept collections, followed by a demonstration on more complex, EHR data.

\subsection{Synthetic Example: Hexagon}
\begin{figure}
\centering
\includegraphics[width = 150px]{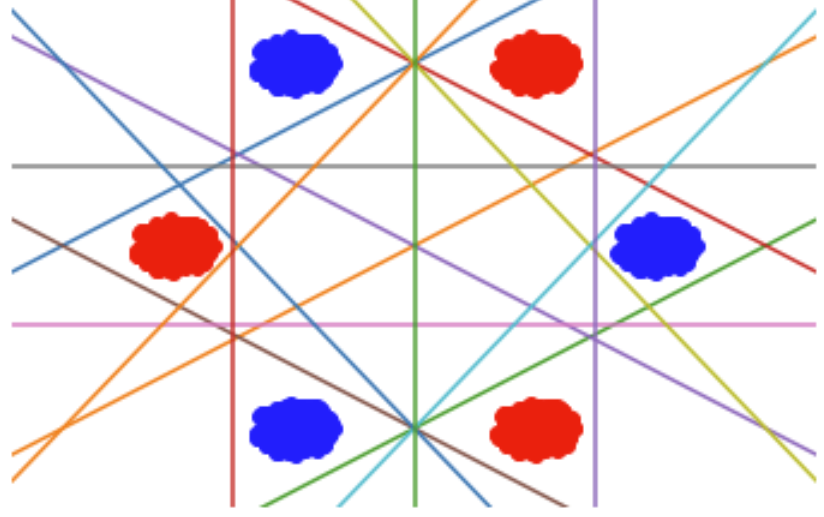}
\caption{The synthetic hexagon  dataset. Each of the 6 clusters contain 200 points. The 15 possible concept decision boundaries between the clusters are denoted by straight lines.}
\label{fig:hexagon}
\end{figure}

We construct a highly non-identifiable, two-dimensional synthetic dataset for a binary classification task, where the concept bottleneck layer $c$ requires a minimum of 3 concepts to classify the data. There are 15 possible binary concepts, which we define as ground truth concepts (each displayed as decision boundaries in Figure \ref{fig:hexagon}) that could be used in conjunction with one another to classify the data. There are 7 possible concept combinations that allow for accurate classification of the dataset.
This dataset is ideal, because we know all possible concept combinations that could explain the data. We can measure how many of these concept combinations are present within our of $M$ proposals. Having more of the combinations present implies that the proposal set covers a larger set of possible explanations for the data. 

\subsection{MIMIC-III}
We use the MIMIC-III dataset to construct a semi-synthetic, early warning dataset  \citep{subbe2001validation}. The input, $x$, includes raw measurements for vitals:  mean blood pressure(arterial), temperature, respiratory rate, heart-rate, and oxygen saturation at 2252 timepoints.
We define 5 synthetic concepts to denote heart-rate, blood pressure, temperature, respiratory rate, and oxygen saturation as being above or below clinically undesirable thresholds. The binary label $y$ denotes whether at least 2 of these thresholds have been breached, indicating urgent need of attention.

\subsection{Evaluation metrics}
We evaluate the methods based on how well they are able to identify the ground truth concepts in the data. We deem that a concept $\mathbf{c}$ was found by a method when one of the $M$ proposals $\mathbf{\tilde{c}}$ matches it with \emph{f1 score} at least 0.9 (on the whole dataset).
For proposals that include multiple concepts that are predictive together, we require that each proposed concept matches a different ground truth concept (or its negation) with f1 score at least 0.9.

\subsection{Implementation Details}
For all of our experiments, we used logistic regression for each input to concept prediction ($x\rightarrow \mathbf{c}$) and for each concept to label prediction ($\mathbf{c}\rightarrow y$). To collect the MCMC samples, we used the reference Hamiltonian Monte-Carlo implementation in Tensorflow \citep{duane1987hybrid, abadi2016tensorflow} with step-size=0.001, number-of-leapfrog-steps=3 and burn-in-steps=1000. For the hexagon dataset, we collect $N=1000$ samples over 10 restarts and for MIMIC-III, we collect $N=100$ samples over 10 restarts.

\section{Results}

\subsection{Hexagon results}

With the hexagon dataset, we are able to compute exactly how many of the 7 possible explanations are found by the different methods. Table \ref{tab:hex_all} shows our results. The best performing selection methods are greedy selection with absolute, percent and euclidean distance metrics. Cosine distance and K-means selection are only able to identify a single valid explanation.

\begin{table}[h]
\caption{Hexagon - Valid explanations (sets of concepts that explain the data) found in the $M=20$ concept-set proposals by different methods.}
\label{tab:hex_all}
\centering
\begin{tabular}{lcccr}
\toprule
Method + Distance metric & Valid explanations  \\
\midrule
Greedy Absolute    & 5/7 \\
Greedy Percent & 6/7 \\
Greedy Euclidean    & 6/7 \\
Greedy Cosine    & 1/7        \\
K-means Euclidean     & 1/7 \\
K-means Cosine      & 1/7 \\
\bottomrule
\end{tabular}
\vspace{-0.2cm}
\end{table}

In the single concept proposal task, we measure how many of the 15 possible single concepts are found by the different methods (Table \ref{tab:hex_single}). Interestingly, we observe that K-means methods slightly outperform greedy methods on this task.

\begin{table}[h]
\caption{Hexagon - Valid concept found in the $M=20$ concept proposals by different methods.}
\label{tab:hex_single}
\centering
\begin{tabular}{lcccr}
\toprule
Method + Distance metric & Valid concepts  \\
\midrule
Greedy Absolute    & 9/15 \\
Greedy Percent & 9/15 \\
Greedy Euclidean    & 9/15 \\
Greedy Cosine    & 9/15  \\
K-means Euclidean     & 11/15 \\
K-means Cosine      & 11/15 \\
\bottomrule
\end{tabular}
\vspace{-0.2cm}
\end{table}

When the human operator identifies a useful concept, it is possible to use our method to generate concepts that complete the explanation. In Table \ref{tab:conditioning} we investigate whether the different methods are able to find the valid explanations conditional on a specified concept. For any given concept, there are two possible explanations that result in a valid explanation. All methods were able to find all possible explanations.

\begin{table}[h]
\caption{Hexagon - Valid completions found in the $M=20$ proposals conditioned on a given concept by different methods.}
\label{tab:conditioning}
\centering
\begin{tabular}{lcccr}
\toprule
Method + Distance metric & Valid completions  \\
\midrule
Greedy Absolute    & 2/2 \\
Greedy Percent & 2/2 \\
Greedy Euclidean    & 2/2 \\
Greedy Cosine    & 2/2  \\
K-means Euclidean     & 2/2 \\
K-means Cosine      & 2/2 \\
\bottomrule
\end{tabular}
\vspace{-0.2cm}
\end{table}

\subsection{MIMIC-III results}

We conduct similar experiments on the MIMIC dataset. In this case, there is only one ground truth explanation, therefore, we measure the number of proposals needed to find this explanation. Table \ref{tab:mimic_m} shows the results. We see that across the board, the methods are able to find the right explanation in their top 3 proposals.

\begin{table}[h]
\caption{MIMIC-III - Minimum number of proposals $M$ needed to find the ground truth explanation.}
\label{tab:mimic_m}
\centering
\begin{tabular}{lcccr}
\toprule
Method + Distance metric & Min $M$  \\
\midrule
Greedy Absolute    & 1 \\
Greedy Percent & 2\\
Greedy Euclidean    & 1 \\
Greedy Cosine    & 1         \\
K-means Euclidean     & 2\\
K-means Cosine      & 3 \\
\bottomrule
\end{tabular}
\vspace{-0.2cm}
\end{table}

Table \ref{tab:mimic_singles} shows the results for single concept proposals. All methods were able to find 4 out of the 5 concepts (missing blood pressure). We observe no significant difference between the different approaches.

\begin{table}[h]
\caption{MIMIC-III - Ground truth concepts found among $M=20$ concept proposals.}
\label{tab:mimic_singles}
\centering
\begin{tabular}{lcccr}
\toprule
Method + Distance metric & Concepts found  \\
\midrule
Greedy Absolute    & 4/5 \\
Greedy Percent & 4/5 \\
Greedy Euclidean    & 4/5 \\
Greedy Cosine    & 4/5 \\
K-means Euclidean     & 4/5 \\
K-means Cosine   & 4/5 \\
\bottomrule
\end{tabular}
\vspace{-0.2cm}
\end{table}

Conditioning on a concept in MIMIC is effective, although it is still unable to find the blood pressure concept. When we fix the respiratory rate concept, we find that the methods are able to find 1 to 2 each of the remaining 4 proposals (Table \ref{tab:mimic_cond}).

\begin{table}[h]
\caption{MIMIC-III - Ground truth concepts found among $M=20$ concept proposals conditional on the respiratory rate concept.}
\label{tab:mimic_cond}
\centering
\begin{tabular}{lcccr}
\toprule
Method + Distance metric & Remaining concepts found  \\
\midrule
Greedy Absolute    & 2/4 \\
Greedy Percent & 2/4 \\
Greedy Euclidean    & 1/4 \\
Greedy Cosine    & 2/4 \\
K-means Euclidean     & 1/4 \\
K-means Cosine   & 1/4 \\
\bottomrule
\end{tabular}
\vspace{-0.2cm}
\end{table}

\section{Discussion}
\label{discussion}

Overall, we found that our method is able to propose diverse concept-based explanations for the data. It is able to find many of the ground truth concepts and also able to propose concepts conditional on selected concepts.

Regarding the differences between methods for constructing the diverse subsets, we found that there are only minor differences between the diversity metrics. There is no clear favourite that we can recommend using.

Between greedy and K-means construction, however, we see that greedy performed slightly better in 3 of our 6 experiments (2 experiments being on-par and one in favour of K-means). We theorize that this is caused by the cluster centers being close concepts that are not along `clear' decision boundaries. It is possible that they are between two ground truth concepts if the cluster contained both of them.

\section{Conclusions}
We investigated methods for generating multiple concept explanations for a dataset that a human expert can choose the most interpretable one from, allowing for recourse when a single explanation offered my a machine learning method is not human understandable.


\section*{Acknowledgements}

This work was funded by a grant from the National Institute of Mental Health (grant no. R01MH123804). The
funders had no role in the design and conduct of the study; collection, management, analysis, and interpretation of the data; preparation, review, or approval of the manuscript; or decision to submit the manuscript for publication.



\end{document}